\documentclass[11pt]{article}
\usepackage[]{acl}
\usepackage{latexsym}
\usepackage[T1]{fontenc}
\usepackage{lmodern}
\usepackage[utf8]{inputenc}
\usepackage{booktabs}
\usepackage{graphicx}
\usepackage{amsmath}
\usepackage{amssymb}
\PassOptionsToPackage{hyphens}{url}
\usepackage{url}
\newcommand{\sw}{Muon-SW}
\newcommand{\bpb}{BPB}

\title{Tune the Baseline First: Most of Muon's Scaled Weight Decay Gain Is a Tuning Gap, but Its Small Real Advantage Grows With Model Size}

\author{
  Esther Xin \\
  Independent Researcher \\
  \texttt{yixin@microsoft.com} \\
  {\small Code and data: \url{https://github.com/ethxin0011/weight_decay_ladder}}
}

\begin{document}
\maketitle

\begin{abstract} Scaled weight decay for Muon (\sw) has been reported to train language models more efficiently than constant decoupled weight decay, but such comparisons usually reuse one learning-rate (LR) and weight-decay (WD) setting for both methods. We re-run the comparison under the MAI-Base-2 protocol: LR and WD are tuned jointly through their product for each method, on a source-scaled mixture-of-experts (MoE) scaling ladder (1.9M to 15M active parameters) that holds tokens per active parameter and data repetition fixed across sizes. We find: \begin{enumerate} \item[(1)] \textbf{Most of the gain is a tuning gap.} At the shared setting \sw{} wins by $0.035$ bits per byte (\bpb), a $1.52\times$ compute saving. After both methods are tuned, the gap is $0.0026$ \bpb{} (95\% CI $[0.0018, 0.0034]$), a $1.03\times$ saving: about 90\% of the advantage disappears. \item[(2)] \textbf{The small advantage that remains is real.} At the smallest size, all four paired seeds favour \sw{} at the tuned settings. \item[(3)] \textbf{That advantage grows with model size.} Across an $8\times$ range of active parameters, the tuned gap rises from $0.0036$ to $0.0051$ \bpb{} while the untuned gap shrinks from $0.035$ to $0.024$, so the share of the advantage that survives tuning rises from about 10\% to 21\%. \item[(4)] \textbf{The best weight decay moves with scale.} The optimal LR$\cdot$WD product grows for both methods, faster for \sw{} (ratio $1.40 \to 1.91$). A law fitted on the two smallest sizes overshoots the third by $1.3$--$1.5\times$, costing up to $0.002$ \bpb. \end{enumerate} The whole study runs on one NVIDIA T4 GPU; we release the code and all run results. \end{abstract}

\section{Introduction}
\label{sec:intro}

New optimizer variants are commonly evaluated by training a model with the
variant and with a baseline, at one shared set of hyperparameters, and
reporting the difference. When the variant changes how strongly a
hyperparameter acts---as a weight-decay schedule does---the shared setting
can be well tuned for one method and poorly tuned for the other, and part
of the reported gain is then a tuning gap. This concern is old
\citep{choi2019empirical,dahl2023algoperf}, but it is sharper for weight
decay, whose effect is coupled to the learning rate
\citep{loshchilov2019adamw,kosson2023rotational,wang2024adamwscale} and
whose optimum moves with model and data size.

Muon \citep{jordan2024muon} is now used for the hidden-layer matrices of
large language models \citep{liu2025muonscalable}, including the 2.8T
parameter MAI-Base-2 \citep{mai2026base2}. \citet{apte2026muonsw} propose
scaling Muon's weight-decay coefficient over training (\sw) and report
gains over constant decay. MAI-Base-2 introduced two practices that make a
cleaner test possible. First, LR and WD are swept jointly on a grid in
(LR, LR$\cdot$WD), because good settings lie on a narrow diagonal of the
(LR, WD) plane and a 2D quadratic fit gives the optimum. Second, models
are compared on a \emph{scaling ladder} with fixed tokens per active
parameter (TPP) and \emph{source scaling}, which keeps the number of
repetitions of every data source fixed across rungs so that small models
do not favour under-regularised settings.

We ask one question: \textbf{how much of \sw's advantage survives once the
baseline is tuned as carefully as the method, and how does that share
change with scale?} Our contributions are:
\begin{itemize}
\item A tuned comparison of \sw{} and constant decay under the MAI-Base-2
LR$\cdot$WD protocol on a source-scaled MoE ladder (three rungs, 1.9M to
15M active parameters, 81 completed runs).
\item The finding that about 90\% of \sw's untuned advantage disappears after
joint tuning at the smallest rung (survival $7.4\%$ in the full 2D
analysis, $10.3\%$ in the 1D slice), but that the residual is consistent
across seeds and grows with scale (survival $10\% \to 13\% \to 21\%$).
\item Measurements of how the optimal LR$\cdot$WD product moves with scale
for each method, a test of three preregistered predictions for the ratio
of optima, and an out-of-sample check of the fitted law at L6.
\item A pre-emption-safe pipeline that runs the full ladder on one T4 GPU.
\end{itemize}

\section{Related work}
\label{sec:related}

\paragraph{Fair optimizer comparison.}
\citet{choi2019empirical} showed that optimizer rankings change once every
optimizer's hyperparameters, including those usually held fixed, are
tuned, and AlgoPerf \citep{dahl2023algoperf} builds tuning budgets into the
benchmark. We apply the same principle to a single optimizer
modification and measure the surviving share explicitly.

\paragraph{Weight decay and scale.}
Decoupled weight decay \citep{loshchilov2019adamw} interacts with the
learning rate through their product, which sets the equilibrium norm and
effective rotation of weights \citep{kosson2023rotational}.
\citet{wang2024adamwscale} derive how WD should change with model and
dataset size for AdamW, and \citet{wortsman2023smallscale} study LR and WD
sensitivity with small proxies. For Muon, \citet{liu2025muonscalable} find
weight decay important at scale; \citet{apte2026muonsw} propose scaling it
over training. MAI-Base-2 \citep{mai2026base2} fits the joint
LR$\cdot$WD optimum as a function of size and TPP.

\paragraph{Scaling ladders and data repetition.}
Scaling laws \citep{kaplan2020scaling,hoffmann2022chinchilla} motivate
comparing methods across sizes rather than at one point. Repeating data
changes what is optimal \citep{muennighoff2023datacon}; source scaling
holds repetition fixed across rungs. Our models use sparse MoE layers
\citep{shazeer2017moe,fedus2021switch}; we also test the
warmup--stable--decay (WSD) schedule \citep{hu2024minicpm}.

\section{Method}
\label{sec:method}

\subsection{Methods compared}
Both methods use Muon for non-embedding 2D parameters and AdamW for all
other parameters, as in MAI-Base-2. Let $\eta_t = \eta\, f(t)$ be the
learning rate at step $t$, where $f(t)\in[0.1, 1]$ is the schedule
(linear warmup, then cosine or WSD decay to $10\%$ of the peak).
\emph{Constant} applies decoupled weight decay with a fixed coefficient
$\lambda$: $\theta \leftarrow \theta\,(1 - \eta_t \lambda)$. \emph{\sw}
\citep{apte2026muonsw} multiplies the coefficient at step $t$ by
$s(t) = \eta_t/\eta_{\max} = f(t)$, so the decay step becomes
$\theta \leftarrow \theta\,(1 - \eta_t^2 \lambda/\eta_{\max})$: decay is
applied at full strength at the peak learning rate and weakens as the
learning rate anneals. Both methods share the same $\eta$ and $\lambda$ at
a given grid point. We write $\bar s$ for the ratio of total decay applied
by \sw{} to that applied by constant decay over a run,
$\bar s = \sum_t f(t)\,s(t) / \sum_t f(t)$; under our cosine schedule
$\bar s = 0.733$ and under WSD $\bar s = 0.956$.

\subsection{Joint LR$\cdot$WD tuning}
Following MAI-Base-2, each rung has a law-predicted learning rate
$\eta_0$ and product $(\eta\lambda)_0$. We sweep multipliers on both: an LR
multiplier and a product multiplier $p$, so a run uses
$\eta = m_\eta \eta_0$ and $\eta\lambda = p\,(\eta\lambda)_0$. At L3 we sweep
the full 2D grid ($m_\eta \in \{0.5, 1, 2\}$, $p$ on a factor-of-2 grid),
fit a 2D quadratic in $(\log_2 m_\eta, \log_2 p)$ per method, and extend
the grid wherever the fitted optimum falls on its edge. At L4 and L6 we
sweep $p$ at $m_\eta = 1$ and fit a 1D quadratic in $\log_2 p$. The same
1D slice is extracted at L3 so that all three rungs can be compared on
equal terms.

\subsection{Quantities reported}
For each rung, the \emph{untuned gap} is the constant-decay \bpb{} minus the
\sw{} \bpb, both at $p=1$; the \emph{tuned gap} is the difference between
the two methods' tuned \bpb. \emph{Survival} is tuned gap divided by
untuned gap. At L3 we also convert gaps into a compute-equivalent speedup
with the loss-versus-compute fit used by our analysis pipeline.

\subsection{Hypotheses for the ratio of optima}
\label{sec:hyp}
We preregistered three predictions for
$r = p^\star_{\rm SW}/p^\star_{\rm const}$. $H_{\rm none}$: the schedule
does not change the optimal product, $r=1$. $H_{\rm int}$: the optimum
matches total decay applied over training, $r = 1/\bar s$ ($1.364$ cosine,
$1.046$ WSD). $H_{\rm end}$: the optimum matches the final-step
coefficient, $r = 10$.

\subsection{Ladder, data and training}
Each rung is named by its layer count $L$, which fixes the hidden size,
head counts and MoE layout (Appendix Table~\ref{tab:ladder}). All rungs
train for 120 tokens per active parameter. The data pool has three public
sources: web text from FineWeb-Edu \citep{penedo2024fineweb}
(\texttt{sample-10BT}; 1{,}174.8M training tokens, 60\% of each batch),
Python code from CodeParrot-clean \citep{codeparrot2022clean} (122.0M
tokens, 25\%) and mathematical web text from OpenWebMath
\citep{paster2023openwebmath} (79.3M tokens, 15\%). Documents shorter than
200 characters and exact duplicates are removed, and 3{,}000 documents per
source are held out for evaluation. We train an 8{,}192-token byte-level
BPE tokenizer on the pool. Under source scaling, web data is seen once and
code and math four times at every rung; lower rungs therefore draw a
proportionally smaller unique subset of each source. The metric is
held-out bits per byte, pooled over the three sources. Main runs use a
cosine schedule; WSD is used for the schedule-robustness check.

\subsection{Compute}
\label{sec:compute}
All runs train on a single NVIDIA T4 (16\,GB) in a low-priority (Spot)
cluster. A self-resubmitting job chain checkpoints every 15 minutes and on
eviction; after a pre-emption a queued successor job resumes from the last
checkpoint. Resumed runs reproduce uninterrupted runs exactly in our
tests. The L6 runs (1.8B tokens each, about 23 GPU-hours) survived 19
evictions without loss. Measured T4 throughput is 29$\times$ that of one
16-core CPU node on L3, so all training was moved to the GPU.

\section{Experimental design}
\label{sec:design}
The study proceeded through preregistered gates. \textbf{Gate 1 (L3,
untuned)} tested whether \sw{} beats constant decay at $p=1$ by more than
seed noise (6 runs). \textbf{Stage 2 (L3 grid)} swept the 2D grid with edge
extensions (39 runs). \textbf{Stage 3 (L3 confirmation)} retrained both
methods at their fitted optima with paired seeds (four seeds per method)
and ran the WSD comparison. \textbf{Stage 4 (L4)} and \textbf{Stage 5 (L6)}
swept $p$ for each method. A planned L9 rung was not run (\S\ref{sec:limits}).

\section{Results}
\label{sec:results}

\begin{table*}[htbp]
\centering
\includegraphics[width=0.92\textwidth]{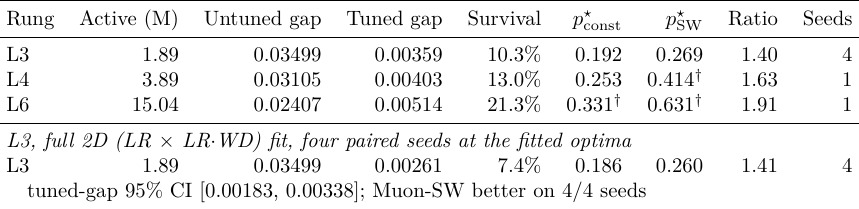}
\caption{Main result. \emph{Untuned gap}: constant-WD BPB minus Muon-SW
BPB, both at the MAI-law setting ($p=1$). \emph{Tuned gap}: the same
difference after each method is tuned over the LR$\cdot$WD multiplier $p$
(quadratic fit in $\log_2 p$). \emph{Survival} = tuned gap / untuned gap.
$p^\star$ is the fitted optimal multiplier; \emph{Ratio} is
$p^\star_{\rm SW}/p^\star_{\rm const}$. Upper block: one-dimensional sweep
at learning-rate multiplier 1 for every rung. Lower block: the full
two-dimensional (LR $\times$ LR$\cdot$WD) L3 analysis, confirmed by four
paired seeds at the fitted optima. $^{\dagger}$Fitted optimum lies outside
the sampled range; the tuned BPB is the better of the fitted and best
observed value. All BPB values are on held-out data.}
\label{tab:main}
\end{table*}

\begin{table}[htbp]
\centering
\includegraphics[width=0.75\linewidth]{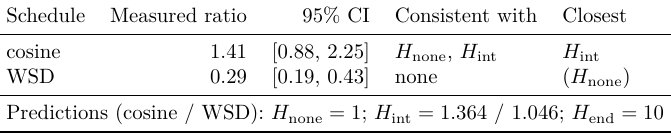}
\caption{Measured ratio of optimal LR$\cdot$WD products (Muon-SW over
constant decay) at L3, with bootstrap 95\% intervals, against the three
preregistered predictions (\S\ref{sec:hyp}). Under the cosine schedule the
data are consistent with no change and with matched integrated decay, and
rule out endpoint matching. Under WSD the ratio falls below every
prediction.}
\label{tab:hyp}
\end{table}

\begin{figure*}[t]
\centering
\includegraphics[width=\textwidth]{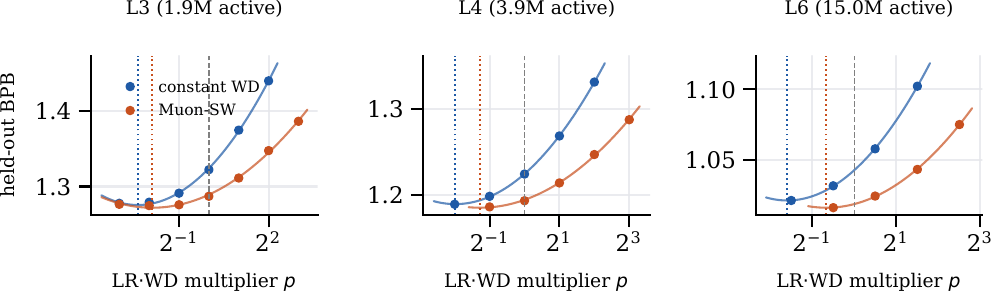}
\caption{Held-out \bpb{} against the LR$\cdot$WD multiplier $p$ (log scale)
for each rung. Points are runs; curves are quadratic fits in $\log_2 p$;
dotted lines mark fitted optima; the dashed line marks the MAI-law setting
$p=1$. At $p=1$ \sw{} (orange) is far better, but constant decay (blue)
recovers most of the gap by moving to a smaller $p$.}
\label{fig:sweeps}
\end{figure*}

\paragraph{At the shared setting, \sw{} is clearly better.}
Gate 1 passed: at $p=1$ on L3 the gap is $0.0350$ \bpb, against a seed
noise of $0.0010$, and it is also positive at TPP 30 ($0.046$). This
reproduces the direction and size of the reported gain and corresponds to
a $1.52\times$ compute-equivalent speedup.

\paragraph{After joint tuning, most of the gap disappears.}
Figure~\ref{fig:sweeps} shows why. Both methods' optima lie below $p=1$,
and constant decay's lies much further below it. Once each is evaluated at
its own optimum, the L3 gap from the full 2D analysis is $0.0026$ \bpb{}
(Table~\ref{tab:main}, lower block), $7.4\%$ of the untuned gap, and the
speedup falls to $1.03\times$. The 1D slice gives a similar $10.3\%$.

\begin{figure}[t]
\centering
\includegraphics[width=0.48\linewidth]{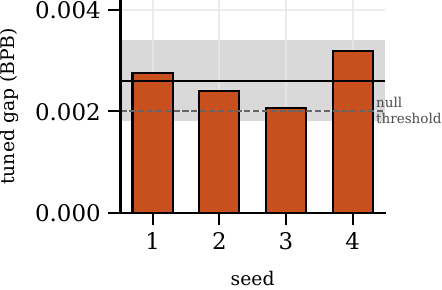}
\caption{Tuned gap at L3 for each of four paired seeds at the 2D-fitted
optima. Black line: mean; grey band: 95\% CI; dashed: the preregistered
$0.002$ \bpb{} threshold.}
\label{fig:seeds}
\end{figure}

\paragraph{The residual is small but real.}
At the fitted optima, all four paired seeds favour \sw{}
(Figure~\ref{fig:seeds}); the 95\% interval $[0.0018, 0.0034]$ excludes
zero. It does not clear our preregistered practical-significance threshold
of $0.002$ \bpb, so we describe the residual as consistent rather than
large.

\begin{figure*}[t]
\centering
\includegraphics[width=0.85\textwidth]{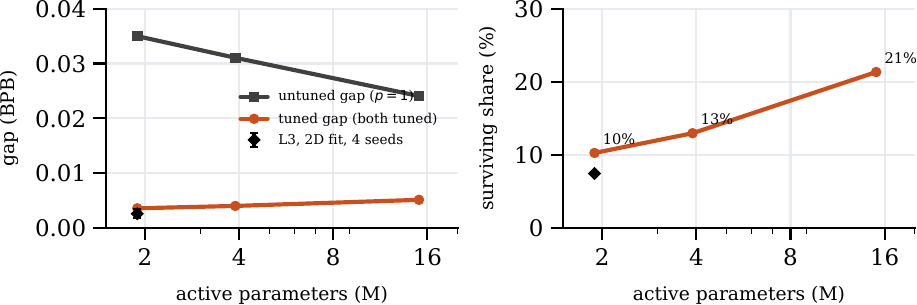}
\caption{Left: untuned and tuned gaps against active parameters; the
diamond is the four-seed L3 estimate with its 95\% CI. Right: share of the
untuned gap that survives tuning.}
\label{fig:scale}
\end{figure*}

\paragraph{The share that survives tuning grows with scale.}
Across rungs (Table~\ref{tab:main}, Figure~\ref{fig:scale}) the untuned gap
shrinks ($0.0350 \to 0.0311 \to 0.0241$) while the tuned gap grows
($0.0036 \to 0.0040 \to 0.0051$). Survival therefore rises from about 10\% to
13\% to 21\%. Both trends are monotone over an $8\times$ range of active
parameters, but the L4 and L6 values come from one seed and from optima at
or just beyond the sweep edge (\S\ref{sec:limits}).

\begin{figure*}[t]
\centering
\includegraphics[width=0.85\textwidth]{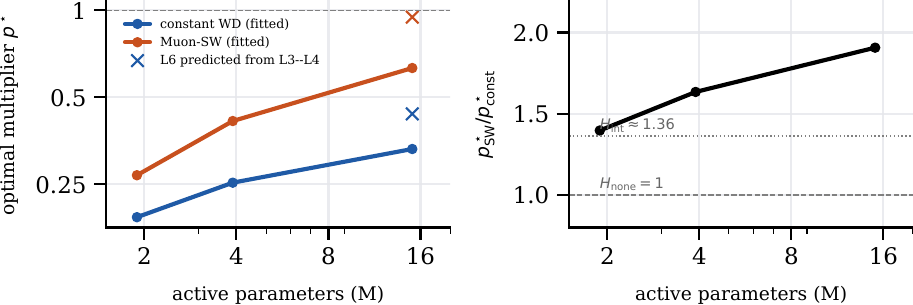}
\caption{Left: fitted optimal LR$\cdot$WD multiplier per method (circles)
and the L6 values predicted from L3 and L4 (crosses). Right: ratio of the
two optima with the $H_{\rm none}$ and $H_{\rm int}$ predictions for the
cosine schedule.}
\label{fig:shift}
\end{figure*}

\paragraph{Optimal decay moves with scale, faster for \sw.}
The fitted optimal multiplier rises with size for both methods
(Figure~\ref{fig:shift}): from $0.19$ to $0.33$ for constant decay and from
$0.27$ to $0.63$ for \sw. Their ratio grows from $1.40$ to $1.91$. A
baseline tuned at small scale and reused at larger scale would therefore
be increasingly mis-tuned, and the direction of that error depends on the
method.

\paragraph{Which hypothesis explains the ratio?}
Under cosine at L3 the ratio is $1.41$ (95\% CI $[0.88, 2.25]$;
Table~\ref{tab:hyp}): consistent with matched integrated decay
($H_{\rm int}=1.36$) and, given its width, with no change; endpoint matching
($r=10$) is ruled out. Under WSD the ratio is $0.29$ ($[0.19, 0.43]$),
below every prediction. We have not resolved this anomaly; it may reflect
an interaction between the long constant-LR phase of WSD and the decay
schedule, and we report it as an open question.

\paragraph{The LR$\cdot$WD law transfers approximately.}
Extrapolating the L3 and L4 optima to L6 predicts multipliers of $0.44$
(constant) and $0.95$ (\sw), against fitted values of $0.33$ and $0.63$
(Appendix Table~\ref{tab:transfer}). Using the predicted value costs
$0.0014$ and $0.0020$ \bpb{} respectively---comparable to the tuned gap
itself. Comparisons at transferred rather than measured optima would
therefore misstate the gap at the scale of interest.

\section{Discussion}
\label{sec:discussion}

\paragraph{What practitioners should take away.}
At small scale, almost all of \sw's reported benefit can be obtained by
tuning the LR$\cdot$WD product of ordinary constant decay. Optimizer
comparisons that hold this product fixed overstate gains roughly tenfold
in our setting. At the same time, the benefit that remains after tuning
grows with scale, which is the regime that matters for large models; our
data are consistent with \sw{} being a small genuine improvement whose
importance increases with size.

\paragraph{Why the untuned gap shrinks.}
The MAI-law setting $p=1$ is too high for both methods on our ladder, but
less so at larger rungs, where both optima move toward $1$. Because
constant decay is penalised more by an over-large product, the untuned gap
mostly measures how far $p=1$ is from constant decay's optimum, and that
distance shrinks with scale.

\section{Limitations}
\label{sec:limits}
\begin{itemize}
\item \textbf{Scale.} Our largest rung has 15M active (94M total)
parameters, far below production models. The planned L9 rung (51M active)
was not run for compute reasons; extrapolating the growth of the residual
beyond L6 is not supported by our data.
\item \textbf{Seeds.} Only L3 has multiple paired seeds. L4 and L6 use one
seed per setting; at L3 the seed standard deviation of the gap is about
$0.0007$ \bpb, small relative to the L4 and L6 tuned gaps, but the
scale trend rests on three points.
\item \textbf{Sweep edges.} The fitted optimum lies outside the sampled
range for \sw{} at L4 and for both methods at L6. We use the better of the
fitted and best observed values; finer sweeps could move the L4 and L6
tuned gaps.
\item \textbf{One dimension above L3.} L4 and L6 sweep only the product at
the law-predicted LR. The L3 1D slice gives a slightly larger survival
($10.3\%$) than the full 2D fit ($7.4\%$), so the 1D numbers may overstate
survival somewhat.
\item \textbf{Schedule.} The WSD result does not match any hypothesis and
was tested only at L3.
\item \textbf{Data.} One data mixture and one tokenizer; results on other
mixtures or with more data repetition may differ.
\end{itemize}

\section{Conclusion}
\label{sec:conclusion}
Most of the advantage reported for Muon's scaled weight decay is a tuning
gap: when constant decay's LR$\cdot$WD product is tuned as carefully as the
method's, about 90\% of the gain disappears at small scale. What remains is
small but consistent, and it grows with model size, from about 10\% to 21\%
of the untuned gap across an $8\times$ range of active parameters, while
the optimal products of the two methods move apart. Optimizer comparisons
should tune the baseline at every scale they report, and claims about
scaled weight decay should be made from tuned comparisons at the target
scale.

\section*{Reproducibility}
Code, run configurations, all 81 run summaries and the script that
regenerates every figure and table (\texttt{make\_figures.py}) are
available at \url{https://github.com/ethxin0011/weight_decay_ladder}.

\bibliography{references}

\clearpage
\appendix
\section{Ladder configuration and transfer check}
\label{app:config}
Table~\ref{tab:ladder} lists the rungs and measured T4 throughput;
Table~\ref{tab:transfer} gives the L6 transfer check.

\begin{table}[htbp]
\centering
\includegraphics[width=0.85\linewidth]{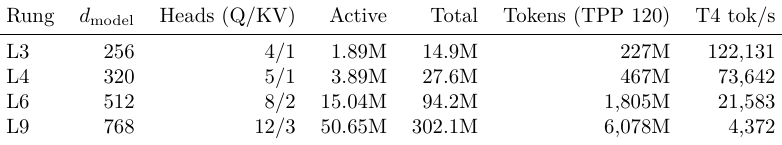}
\caption{Scaling-ladder rungs. Each rung is defined by its layer count
$L$, which fixes the rest of the architecture. Tokens are for 120
training tokens per active parameter. Throughput is measured on one
NVIDIA T4 (\S\ref{sec:compute}). L9 was planned but not trained.}
\label{tab:ladder}
\end{table}

\begin{table}[htbp]
\centering
\includegraphics[width=0.75\linewidth]{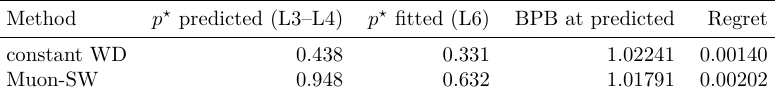}
\caption{Transfer of the LR$\cdot$WD law to L6. The optimal multiplier
predicted by extrapolating the L3 and L4 optima overshoots the fitted L6
optimum by $1.3$--$1.5\times$ for both methods. Regret is the BPB lost by
using the predicted rather than the fitted value.}
\label{tab:transfer}
\end{table}

\end{document}